\documentclass[11pt,a4paper]{article}

\usepackage{amsmath,graphicx,fullpage}
\usepackage{cite}
\usepackage{booktabs}
\usepackage{multirow}
\usepackage{array}
\usepackage{enumerate}
\usepackage{color}
\usepackage{url}
\usepackage{pdflscape}
\newcolumntype{x}[1]{%
>{\centering\hspace{0pt}}p{#1}}%

\usepackage{fancyhdr}
\date{}
\title{\Large {\bf Atmospheric Turbulence Removal with Complex-Valued Convolutional Neural Network}}
\author{Nantheera Anantrasirichai\footnote{This work was funded by the UKRI
MyWorld Strength in Places Programme (SIPF00006/1).} \\ University of Bristol}

\begin{document}

\maketitle

\begin{abstract}
Atmospheric turbulence distorts visual imagery and is always problematic for information interpretation by both human and machine. Most well-developed approaches to remove atmospheric turbulence distortion are model-based. However, these methods require high computation and large memory making real-time operation infeasible. Deep learning-based approaches have hence gained more attention but currently work efficiently only on static scenes. This paper presents a novel learning-based framework offering short temporal spanning to support dynamic scenes. We exploit complex-valued convolutions as phase information, altered by atmospheric turbulence, is captured better than using ordinary real-valued convolutions. Two concatenated modules are proposed. The first module aims to remove geometric distortions and, if enough memory, the second module is applied to refine micro details of the videos. 
Experimental results show that our proposed framework efficiently mitigates the atmospheric turbulence distortion and significantly outperforms existing methods.
\end{abstract}

\section{Introduction}
\label{sec:intro}

Light propagating through atmospheric turbulence, caused by heat sources at different temperatures, appears as a combination of blur, ripple and intensity fluctuations in video sequences. This phenomenon obviously deteriorates the visual quality and the performance of computer vision techniques, including face detection, object tracking, 3D rendering, etc. 
Mitigating the atmospheric turbulence distortion has been attempted both online and offline. Online systems are integrated into optical devices to operate in real time which limits visual quality, whilst the offline approach aims for better restoration results, but it could be slower, even using high performance computing machines.

Mathematically, the model of atmospheric turbulence effects is generally expressed as $y = Dx + n$, where $x$ and $y$ are the ideal and observed images respectively, $D$ represents unknown geometric distortion and blurs from the system, and $n$ represents noise. Despite being simple, this problem is irreversible thereby resulting in an imperfect solution in practice.  Traditional methods  have solved this problem by modelling it as a point spread function (PSF) and then employing blind deconvolution with an iterative process to estimate $x$ \cite{Harmeling:online:2009}. Alternatively, image fusion techniques, where  only good information among frames are selected to reconstruct a new clearer frame, achieve desirable restoration performances \cite{Anantrasirichai:Mitigating:2012}. However, there are  two major problems with these model-based methods: i) high computational complexity, meaning that a real-time implementation is almost impossible, ii) artefacts from moving objects due to imperfect alignment created when combining multiple images \cite{Anantrasirichai:Atmospheric:2018}.

Deep learning has emerged as a powerful tool to find patterns, analyse information, and to predict future events. These capabilities make the learning-based techniques widely used in image and video applications \cite{Anantrasirichai:Artificial:2021}. For atmospheric turbulence removal, deep learning is still in the early stages and all proposed methods are based on convolutional neural networks (CNNs). The first deep learning-based method, proposed by Gao at al., \cite{Gao:Atmospheric:2019}, follows the assumption that the spatial displacement between frames due to atmospheric turbulence has a Gaussian distribution. The state-of-the-art Gaussian denoiser, DnCNN \cite{Zhang:dncnn:2017}, architecture is hence used. Later, the method proposed in \cite{Mao:accelaring:2021} employed UNet architecture, which was originally introduced for medical image segmentation \cite{Ronneberger:Unet:2015}. They adjusted the input channels of the UNet to accept a 50-frame concatenated volume  and the output was a single restored frame. As this method requires a long length of inputs, it is infeasible to restore the distorted videos with moving objects.
Vinta et al. \cite{Vint:analysis:2020} have investigated the performance of mitigating atmospheric turbulence effects with various state-of-the-art architectures, originally proposed for denoising, deblurring, and super-resolution. The results from their report are very promising. However, they studied only on synthetic static scenes. 

Some proposed methods, that involve supervised learning, imitate a traditional workflow of the model-based techniques. Firstly, the reference frame is constructed. A series of frames is subsequently registered to this reference frame, and then the registered frames are averaged. Finally, a deep learning-based deblurring technique, e.g. DnCNN \cite{Nieuwenhuizen:Deep:2021} and pyramid fusion network \cite{Fazlali:Atmospheric:2022}, is applied to sharpen the averaged frame. Obviously they are not end-to-end deep learning-based frameworks and the computational speed is not reduced. More recent methods proposed end-to-end deep learning architectures. A WGAN (Wasserstein generative adversarial network) is employed in \cite{Chak:Subsampled:2021}, where the multiple lucky frames are fed into the UNet generator. This however appears only to work well for static scenes. A framework in \cite{Wang:deep:2021} comprises two CNNs: i) phase aberration correction and ii) Zernike coefficients reconstruction. However, only one small image result is reported, so its performance cannot be commented upon. As ground truth is unavailable for the atmospheric turbulence problem, a self-supervised technique has been proposed in \cite{Li:Unsupervised:2021}, where geometric distortion is removed using a grid-based rendering network. The method estimates spatial displacements between the distorted frames. The clean frame is consequently the output when the zero-displacement map is fed. This method however requires deblurring as post-processing and also needs ten input frames, so it may not work properly if moving objects are present.

In this paper, we aim to exploit deep learning mechanism to mitigate atmospheric turbulence distortion in the video sequences. We restrict our framework to require a small number of inputs so that i) it can operate in real time, and ii) it will not create artefacts due to unachievable alignment through multiple convolutional layers, particularly when fast moving objects are present. Restoring the video is done in a temporal sliding window.
Our framework comprises two modules, aiming to remove geometric distortion and to enhance visual appearance, called a \textit{distortion mitigating module} and a \textit{refinement module}, respectively. Our networks operate in the complex domain, allowing richer representation of phase than in the real domain. This is directly inspired by our previous work \cite{Anantrasirichai:Mitigating:2012, Anantrasirichai:Atmospheric:2018, Anantrasirichai:Atmospheric:2013}, where the process is done in the complex wavelet domain. Additionally, deep complex networks \cite{Trabelsi:deep:2018} have proved superior performance in image recognition over the real-valued only networks. We describe our version of a complex-valued CNN in Section \ref{sec:complexcnn}, and our network architecture in Section \ref{sec:proposed}. 

As stated above, the restoration of atmospheric turbulence distortion is a ill-posed problem. Ground truth is generally not available. We then intensively test our framework that is trained with a combination of the synthetic and the real datasets, where pseudo ground truth is generated for the real dataset. In Section \ref{sec:results}, the training and test datasets are described, followed by the experimental results and discussion. Finally the conclusion of this work is present in Section \ref{sec:conclusion}.

\section{Mitigating turbulent distortion with complex-valued convolutional neural networks}
\label{sec:complexcnn}

\subsection{Why complex values?}

Turbulent medium causes phase fluctuations \cite{Kolmogorov:local:1991}. This is exhibited in the image as a phase shift in the Fourier domain and in the wavelet domain, and the amount depends approximately linearly on displacement \cite{Chen:registation:2011, Hill:Undecimated:2015}. Following quasi-periodic property, the ripple effect of the atmospheric turbulence causes micro displacement between frames with random amount and direction \cite{Li:Suppressing:2009}. The phase of each pixel is consequently altered randomly, whilst the magnitude of high frequency may be decreased due to mixing signals, leading to blur on the image. 
Many simulated methods also model atmospheric turbulence in phase domain. For example, authors in \cite{Woods:Lucky:2009} describe atmospheric turbulence effects with wavefront sensing principles, and employ phase diversity to estimate severity level. Chimitt et. at. \cite{Chimitt:Simulating:2020} model phase distortion as a function of frequency, focal length, aperture diameter and a random vector. They show that the phase distortion introduces a random tilt to the PSF.
With the above reasons, we therefore employ complex-valued convolution to extract phase fluctuations and atmospheric turbulence distortion is removed in both real and imaginary components.

\subsection{Complex-valued convolutional layer (CConv)}
Similar to previous work \cite{Trabelsi:deep:2018, Cole:Analysis:2021}, implementing complex-valued convolution is straightforward. We define that a feature value $I$ in the feature map has a complex value as $I = I_{\Re} + i I_{\Im}$, where $i=\sqrt{-1}$ is the imaginary unit, $I_{\Re} = \Re\{I\}$ and $I_{\Im} = \Im\{I\}$ are the real and the imaginary components, respectively. With the distributive property, the convolution ($*$) of $I$ and a complex-valued kernel $H = H_{\Re} + i H_{\Im}$ is then expressed as 
\begin{equation}
\label{eq:complexconv}
\begin{split}
    I * H &=  (H_{\Re} + i H_{\Im})*(I_{\Re} + i I_{\Im})  \\
    &= (H_{\Re}*I_{\Re}  - H_{\Im}*I_{\Im} ) +  i (H_{\Im}*I_{\Re} +  H_{\Re}*I_{\Im}).
\end{split}
\end{equation}
\noindent Eq. \ref{eq:complexconv} can be straightforwardly implemented with four separate convolutions using existing tools, e.g. \texttt{torch.nn.Conv2d} in PyTorch. We do not use a pooling layer, but down-sampling feature maps is done through the transposed convolution with a stride of 2 (CConvTran).

\subsection{Complex-valued activation function} 
For an activation function, we select the rectified linear unit function, ReLU, because of its simplicity. The experiment in \cite{Cole:Analysis:2021} shows that applying the ReLU function  to the real and imaginary parts separately achieves better image reconstruction than applying to the magnitudes alone. Therefore, our complex ReLU function, CReLU, is defined as
$\text{CReLU}(I) = \text{ReLU}(I_\Re) + i \text{ReLU}(I_\Im)$.

For more stable training, we employ the leaky ReLU activation function. This prevents the ‘dying ReLU’ problem, where zero gradients happen when spikes of high frequencies due to spatially turbulent variation amongst frames occur in several training batches consecutively. The leaky ReLU allows a small gradient when the unit is not active (negative inputs) so that the backpropogation will always update the weights. We set the gain $\alpha$ of the leaky ReLU for the negative values to 0.2, and our complex-valued  leaky ReLU (CLReLU) is defined as

\begin{equation}
 \text{CLReLU}(I) = \text{LReLU}(I_\Re) + i \text{LReLU}(I_\Im).
\end{equation}

 \subsection{Batch normalisation} 
 
Batch normalisation generally improves stability of the training process. We have tried to normalise the feature maps using the method proposed in \cite{Trabelsi:deep:2018} and using a batch norm function for real values \cite{Ioffe:batch:2015}. With both techniques, the colours of results become flattened. So, we have decided not to create a new normalisation for complex-valued tensors, nor apply any normalisation process within our complex-valued neural network as used in the applications of classification and Gaussian denoising. Instead, we scale and shift the input images to $[-1, 1]$ and utilise the element-wise hyperbolic tangent function (\texttt{Tanh}) to cap the output to $[-1, 1]$.




\section{Proposed architecture}
\label{sec:proposed}

\begin{figure} [t!]
    \centering
    \includegraphics[width=0.65\textwidth]{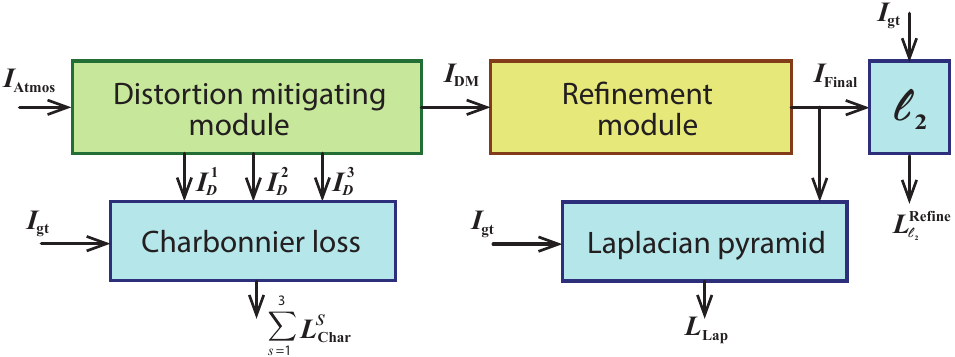}
    \caption{Diagram of the proposed framework with loss functions for training. The input $I_\text{Atmos}$ is a group of distorted frames, $I_\text{gt}$ is a clean current frame. }
    \label{fig:diagramall}
\end{figure}

The proposed end-to-end framework comprises two modules as shown in Fig. \ref{fig:diagramall}. The input is a group of distorted frames, $I_\text{Atmos} = \{I^{t+n}\}_{n\in [-N_b, N_f] }$, where $t$ is the current timestamp, $N_f$ and $N_b$ are the numbers of forward and backward frames, respectively. The output is the restored version of the current frame, $I_\text{Final}$. We call the first one as a distortion mitigating module, where the geometric distortion due to atmospheric turbulence is removed, giving the output $I_\text{DM}$ (described in Section \ref{ssec:distortion}). The second module is for detail refinement, called a refinement module (described in Section \ref{ssec:refine}). This second module is to ensure the sub-details in feature and spatial spaces are as close as those of the ground truth.

\subsection{Distortion mitigating module}
\label{ssec:distortion}

\begin{figure} [t!]
    \centering
    \includegraphics[width=0.7\columnwidth]{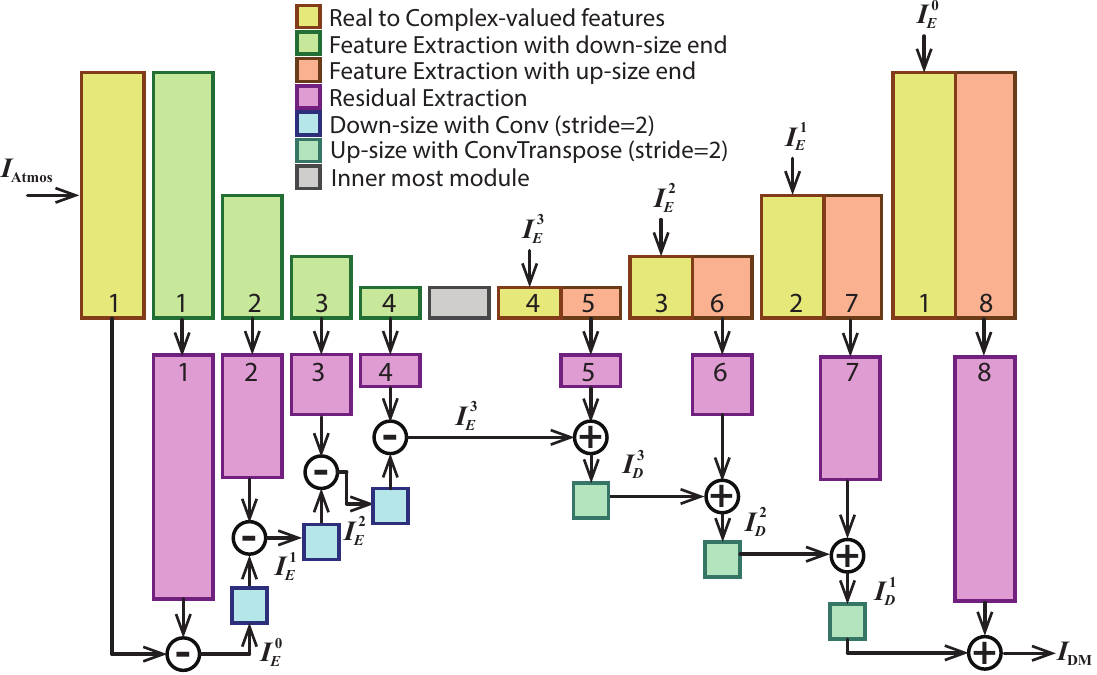}
    \caption{Distortion mitigating module combining use of an encoder to extract distorted features and a decoder to reconstruct distortion-free frame. The distorted-suppressed frames $I_E^s$ in several resolutions at the encoder are bridged with the decoder by concatenation. At the decoder the edge features are added up to create sharper frame $I_D^s$. The number shown at each sub-module is associated with the name listed in Table \ref{tab:config}}.
    \label{fig:diagram}
\end{figure}

The diagram of the proposed network is illustrated in Fig. \ref{fig:diagram} and the network configuration details are listed in Table \ref{tab:config}. The number of input frames is $N_t = N_b+N_f+1$. The number of output frames is also $N_t$ if the refinement module is enabled; otherwise the number of the output is one, which is the current frame.
Our distortion mitigating module is an encoder-decoder architecture with the connection between distorted-free features restored at the encoder and the decoder (appearing as $I_E^s$ in Fig. \ref{fig:diagram}, where $s$ is the resolution level with the maximum level of $S$). The encoder part estimates geometric distortion at different resolution level. Each feature extraction module (green block) comprises nine $3\times3$ CConvs, and each CConv is followed with a CLReLU. We reduce and increase the size of the feature maps with a 4$\times$4 convolution and a 4$\times$4 CConvTran, respectively, with a stride of 2. The output of each feature extraction module is 64 feature maps, $\{I^c\}_{c \in [0, 63]}$.

We include the residual extraction sub-modules (pink block) to convert the distortion represented in the complex-valued feature space to the image space.
The output of the residual extraction module is subtracted from the distorted inputs. The complex-valued output is converted back to real-valued image, producing distortion-free images with real values $I_E^s = |I^c|^s$. 

At the decoder, the features of the $I_E^s$ are extracted and concatenated to the features decoded from the inner-most module (grey block). Then the residual extraction sub-modules are exploited again to extract the high frequencies, such as edges and structures, to add in $I_D^s$ to enhance sharpness of the final image $I_\text{DM}$. This is done in the pyramid manner which has proved its success in image restoration not only when using CNNs \cite{Lai:Fast:2019}, but also wavelet-based reconstruction \cite{Anantrasirichai:Atmospheric:2013}. 

\begin{table}[t!]
	\centering
	\caption{Configuration of the distortion mitigating module. Dimension of the feature map is height$\times$width$\times$channels in complex-valued format (one channel has real and imaginary components). $N_t$ is the number of frames. Colour videos are used here. We simply concatenate R, G and B, resulting total 3$N_f$ channels. If not indicated, the convolution is applied with a stride (st) of 1.}
	\footnotesize
		\begin{tabular}{ccc}
		\hline
		 Sub-module & layers & output dimension \\
			\hline
			input & & 256$\times$256$\times 3 N_t$ \\ \hline
			real2complex1,4 &  & 256$\times$256$\times$64\\
			real2complex2 & (3$\times$3 Conv, CLReLU) & 128$\times$128$\times$64\\ 
			real2complex3 &  & 64$\times$64$\times$64\\
			real2complex4 &  & 32$\times$32$\times$64\\\hline
			feature1 &   & 256$\times$256$\times$64\\
			feature2 & 9$\times$(3$\times$3 CConv, CLReLU)$+$ & 128$\times$128$\times$64\\
			feature3 & (4$\times$4 CConv st=2, CLReLU) & 64$\times$64$\times$64\\
			feature4 & & 32$\times$32$\times$64\\ \hline
			\multirow{2}{*}{inner most} & (3$\times$3 CConv, CLReLU) & 16$\times$16$\times$64\\
			& (4$\times$4 CConvTran st=2, CLReLU) &  32$\times$32$\times$64\\ \hline
			feature5 & 9$\times$(3$\times$3 CConv, CLReLU)$+$  & 64$\times$64$\times$64\\
			feature6 & (4$\times$4 CConvTran st=2, & 128$\times$128$\times$64\\
			feature7 &  CLReLU) & 256$\times$256$\times$64\\ \hline
			feature8 & 9$\times$(3$\times$3 CConv, CLReLU) & 256$\times$256$\times$64\\ \hline
			residual1,8 & \multirow{4}{*}{(3$\times$3 CConv, LReLU)} & 256$\times$256$\times 3 N_t$ \\
			residual2,7 & & 128$\times$128$\times 3 N_t$ \\
			residual3,6 & & 64$\times$64$\times 3 N_t$ \\
			residual4,5 & & 32$\times$32$\times 3 N_t$ \\
				\hline 
		 \multicolumn{3}{l}{Conv: Convolutional layer, CConv: Complex-valued convolutional layer} \\
		 \multicolumn{3}{l}{CConvTran: Complex-valued  transposed convolutional layer, st: stride} \\
		 \multicolumn{3}{l}{LReLU: Leaky ReLU, CLReLU: Complex-valued Leaky ReLU}
		\end{tabular}
	\label{tab:config}
\end{table}

\subsection{Refinement module}
\label{ssec:refine}

The first module generally reduces geometric distortion significantly. However, when facing strong atmospheric turbulence, the remaining effect resulting from spatial variation of an unknown PSF requires a deblurring process. We simply adapt the UNet architecture \cite{Ronneberger:Unet:2015} with complex-valued convolution to deal with this. We set the depth of our complex-valued UNet to 5, which is reasonable for the input size of 256$\times$256 pixels. The real-to-complex sub-module, similar to that used in the distortion mitigating module, is attached before the complex-valued UNet. At the end, the complex-to-real sub-module is added to give the final output $I_\text{Final}$.

\subsection{Loss functions}

Our training loss, $L_{\text{Train}}$ is computed using a combination of three loss functions, Charbonnier loss, Laplacian pyramid loss \cite{Bojanowski:Optimizing:2018} and $\ell_2$, as shown in Eq. \ref{eq:lostfn}. In the distortion mitigating module, the pixel-wise loss due to the atmospheric turbulence effect might create some outliers. The Charbonnier loss is therefore used because it combines the benefits of $\ell_1$ and $\ell_2$ appearing to handle outliers better \cite{Lai:Fast:2019}. It is defined as $L_\text{Char}(x)=\sqrt{x^2 + \epsilon^2}$, $\epsilon=1\times10^{-3}$. This loss captures the content similarly between the reconstructed images and the ground truth at different resolution levels, $s \in S$. Here we employ 4 resolution levels, i.e. $S\in[0,...,3]$, $I_D^0 = I_\text{DM}$ (see Fig. \ref{fig:diagram}). 

 Laplacian pyramid loss $L_\text{Lap}$ is applied at the final output of the refinement module. This is to ensure that image structures presenting at different scales are similar to those of the ground truth. Following \cite{Bojanowski:Optimizing:2018}, $L_\text{Lap}=\sum_j 2^{2j} | \Lambda^j (I_\text{Final}) - \Lambda^j(I_\text{gt})|_1$, where $\Lambda^j (x)$ is the $j$-th level of the Laplacian pyramid representation of $x$. The output of the refinement module will no longer suffer from the atmospheric turbulence distortion, leading to significantly fewer outliers compared to the output of the first module. Therefore we include a mean-square-error loss, $L^\text{Refine}_{\ell_2}$, instead of the Charbonnier loss, to maximise content and colour similarities to the ground truth.
 
\begin{equation}
\label{eq:lostfn}
    L_{\text{Train}} =  \sum_s^S L^s_\text{Char} + L_\text{Lap} + L^\text{Refine}_{\ell_2}
\end{equation}

\section{Experiments and discussion}
\label{sec:results}

We trained and tested our proposed framework with both synthetic and real datasets, and then compared the performance with some related work. Most existing methods are however developed for static scenes \cite{Nieuwenhuizen:Deep:2021, Fazlali:Atmospheric:2022,Chak:Subsampled:2021,Wang:deep:2021}, and some of them are not truly end-to-end deep learning frameworks \cite{Nieuwenhuizen:Deep:2021, Fazlali:Atmospheric:2022}. Also, their source code is not available at the time of writing. Therefore, we compared the performance of our method with states of the art of the image and video restoration: i) UNet \cite{Ronneberger:Unet:2015}, the backbone architecture of many image denoisers and restorers \cite{Abdelhamed:NTIRE:2020}, ii) EDVR \cite{Wang:EDVR:2019}, the winning solution in NTIRE19 Challenges on video restoration, and iii) FFDNet \cite{Zhang:FFDNet:2018} offering the best denoising performance reported in many surveys \cite{Anantrasirichai:Artificial:2021, Tian:Deep:2020}. 

For fair comparison, all methods were modified to accept multiple frames, and were retrained with our datasets. If not indicated, the input $N_t$ was 5 consecutive frames ($N_b$=$N_f$=2) and the model was trained using a temporal sliding window procedure for 200 epochs. Adam optimizer was employed with an initial learning rate of 0.0001. All experiments were carried out using the computational facilities of the Advanced Computing Research Centre, University of Bristol (\url{http://www.bristol.ac.uk/acrc/}).


\subsection{Datasets}
\label{ssec:Datasets}
The main challenge of the atmospheric turbulence restoration is lack of ground truth in the real-world scenarios. Some existing datasets provide  clean ground truth data, but they are all for static scenes \cite{Anantrasirichai:Atmospheric:2013, Hirsch:Efficient:2010}. The Open Turbulent Image Set (OTIS) \cite{Gilles:Open:2017} is the only dataset captured with dynamic scenes. However, its primary aim is object tracking, so the ground truth is only a bounding box around a moving toy car, rather than a clean sequence.
We therefore generated some synthetic datasets, used the pseudo ground truth for the real datasets, and then trained the models using a combination of these.

Our real dataset contains 14 paired videos, available at \url{https://bit.ly/CLEAR_datasets}, ranging from 100 frames to 4,000 frames with different resolutions. The longer videos include panning and zooming. The pseudo ground truth of the real datasets was generated off-line using the CLEAR method \cite{Anantrasirichai:Atmospheric:2018}. 
For the synthetic data, we created seven 480$\times$800 atmospheric turbulence sequences using nine predefined PSFs of atmospheric turbulence provided in \cite{Hirsch:Efficient:2010}. The process was performed on a frame-by-frame basis. The spatially variant blur was created by applying a randomly selected PSF to different parts of the image, and for each image the PSFs were resized randomly so that the strength of ripple effects and degree of blur vary between frames. Finally, a Gaussian noise with zero mean and random variance was added. 
These synthetic datasets are available at \url{https://bit.ly/Synthetic_datasets}.




\subsection{Synthetic examples}

The experiments on the synthetic examples aim to perform objective quality assessment as the ground truth is known. We exploit PSNR and SSIM applied to each frame in the sequence and the results shown are the average of all test frames. If the videos are in colour, we compute PSNR and SSIM for each colour channel separately and the results are averaged. The results shown in Table \ref{tab:synresult} reveal that the proposed method achieves the best performance, which improves the visual quality by 20\% and 7\% from the raw distorted sequences in terms of PSNR and SSIM, respectively.

\begin{table*}[t!]
	\centering
	\caption{Average objective quality of the restored synthetic and real sequences. Bold and underline indicate the best and the second best performances.  }
	\footnotesize
		\begin{tabular}{ccccccc}
		\hline
		 Method & Raw & UNet \cite{Ronneberger:Unet:2015} & EDVR \cite{Wang:EDVR:2019} &  FFDNet \cite{Zhang:FFDNet:2018} & Proposed \\
			\hline
			\multicolumn{6}{c}{Synthetic data} \\
			PSNR & 28.80 & \underline{34.197} & 33.365 & 29.852 & \textbf{34.533} \\
			SSIM & 0.902 & \underline{0.959} & 0.953 & 0.950 & \textbf{0.961} \\
		\hline 
		\multicolumn{6}{c}{Real data with pseudo ground truth} \\
			PSNR & 28.18 & \underline{33.764} & 33.667 & 26.091 & \textbf{34.310}\\
			SSIM & 0.884 & 0.931 & \underline{0.936} & 0.901 & \textbf{0.938} \\
		\hline
		\end{tabular}
	\label{tab:synresult}
\end{table*}

\subsection{Real atmospheric turbulence sequences}

\begin{figure}
    \centering
    \includegraphics[width=0.77\columnwidth]{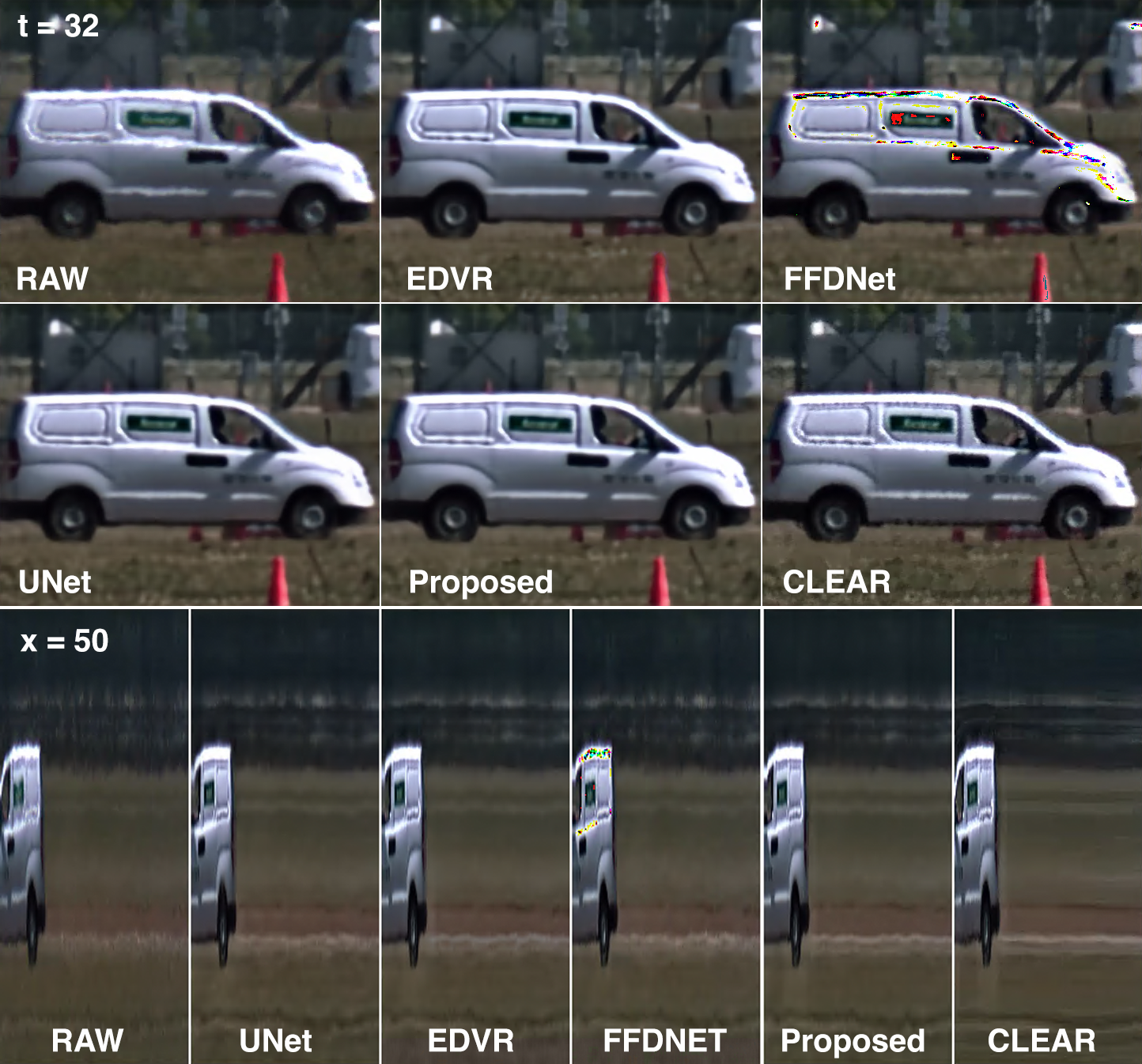}
    \caption{Subjective comparison of `Van' sequences. Top-Middle: x-y frame at t=32. Bottom: y-t plane at x = 50. The CLEAR results are used as pseudo ground truth. }
    \label{fig:van_x50}
\end{figure}

\begin{figure}
    \centering
    \includegraphics[width=0.9\columnwidth]{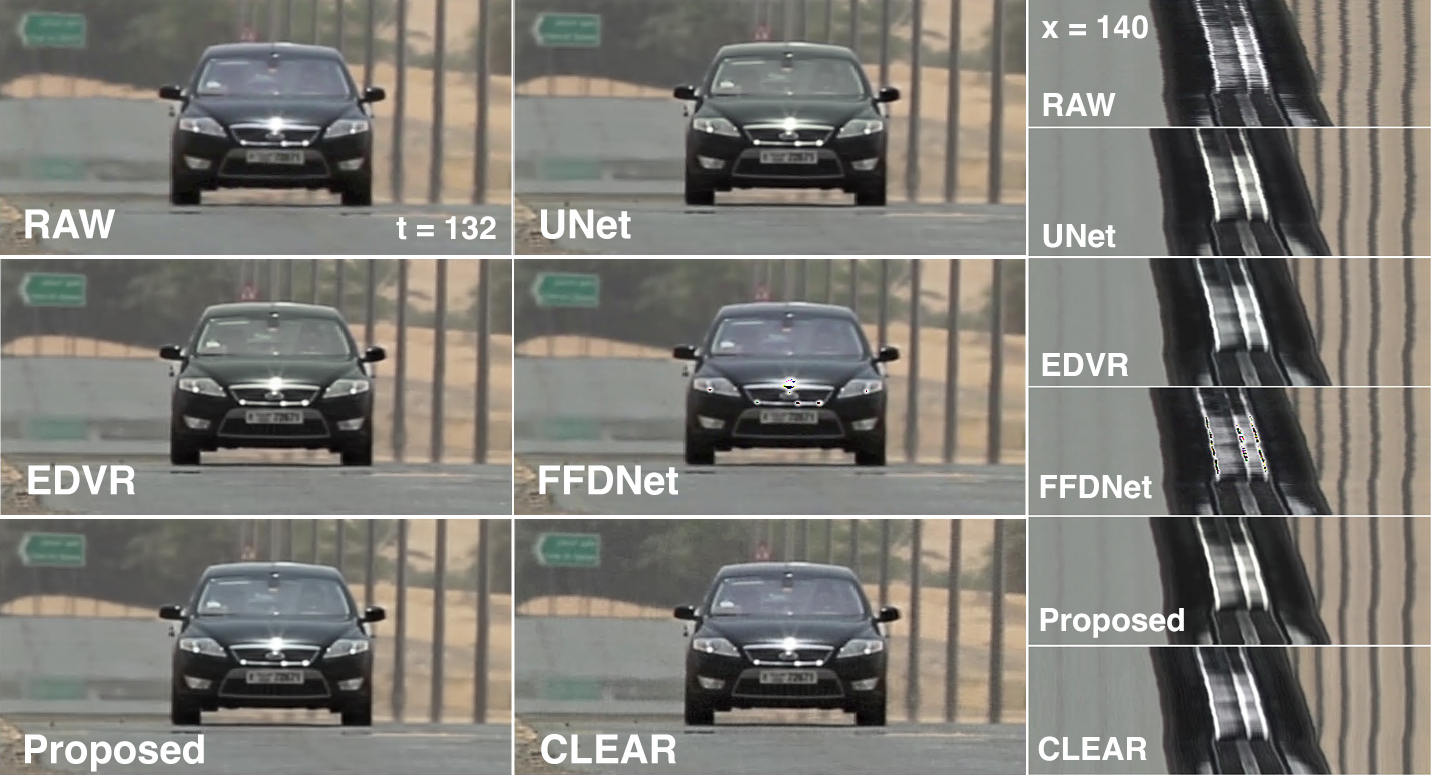}
    \caption{Subjective comparison of `Car' sequences. Top-Middle: x-y frame at t=132. Bottom: x-t plane at y = 140. The CLEAR results are used as pseudo ground truth.}
    \label{fig:moving_car_t132}
\end{figure}

The objective quality assessment was done by comparing the output with the pseudo ground truth and the results are shown in Table \ref{tab:synresult}. This might not represent a true performance comparison, but it can be thought of as how well the methods can remove the temporal variation, since the CLEAR method achieves desirable mitigation of the atmospheric turbulence effect (see the y-t and x-t planes of CLEAR in Fig. \ref{fig:van_x50} and Fig. \ref{fig:moving_car_t132}).

The subjective results are shown in Fig. \ref{fig:van_x50} and \ref{fig:moving_car_t132} (zoom in for better visualisation and comparison). The restored frames reveal how smooth the edges are recovered and how much detail is restored. These CNN-based methods produce smoother straight lines than the pseudo ground truth generated from CLEAR, but the edges are not as sharp. We also show the cross sections of the video sequences in x-t or y-t plane to demonstrate how much ripple effect is suppressed. The proposed method clearly produces better visual quality than other methods. EDVR performs well only when the atmospheric turbulence distortion is subtle, like the `car' sequence in Fig. \ref{fig:moving_car_t132}. FFDNet cannot reduce the ripple effect and some signals are clipped.

\subsection{Ablation study}

\subsubsection{Real-valued vs complex-valued convolution}
\begin{figure}
    \centering
    \includegraphics[width=0.7\columnwidth]{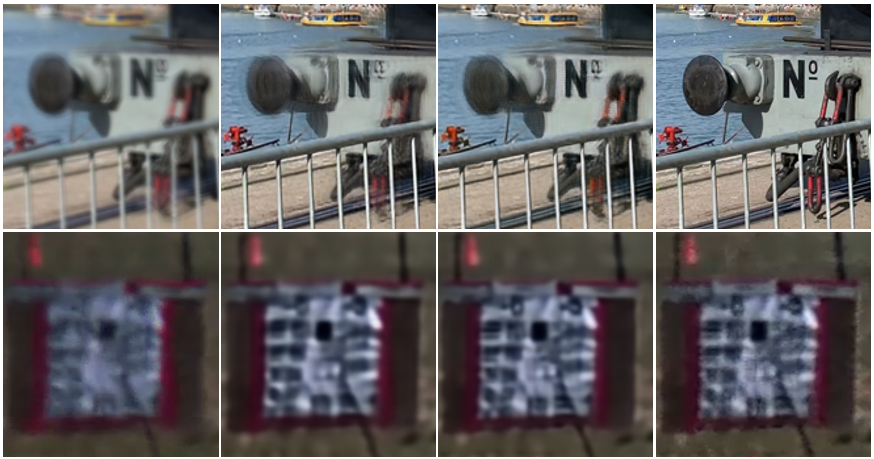}
    \vspace{-2mm}
    \begin{tabular}{x{2.25cm}x{2.25cm}x{2.25cm}x{2.25cm}}
\footnotesize Distorted & \footnotesize  Real & \footnotesize  Complex & \footnotesize  Ground truth\\ 
\end{tabular} 
    \caption{Result comparison of real-valued vs complex-valued convolution. Top-row: Cropped frame 10 of the synthetic sequence `Train'. Bottom-row: Cropped frame 120 of the real sequence `Van'. The ground truth of the `Van' sequence is pseudo. Please zoom in for better visualisation.}
    \label{fig:compareRealComplex}
\end{figure}

We studied the effect of using complex values by comparing the performance of the traditional UNet with the complex-valued UNet. Some results of the synthetic and real datasets are shown in Fig. \ref{fig:compareRealComplex}. As expected, the complex-valued UNet can better remove the ripple effect, and produces better temporal consistency than the real-valued UNet. This can be seen in the area near the letter `N' of the `Train' sequence, where motion artefacts are clearly present.  Additionally, the restored results of the complex-valued UNet appear to be sharper than those of the real-valued one. 

The complex-valued convolutions require more memory and training time than the conventional convolutions. On the 640$\times$360 sequences, the complex-valued UNet generates the restored results approximately 50 frames/sec, compared to about 60 frames/sec for the real-valued UNet.
Another drawback of the complex-valued convolution we found is that sometimes the model loses one colour channel which consequently requires more training iterations to recover it, or restarting the training process.

\subsubsection{With vs without refinement module}

\begin{figure}
    \centering
    \includegraphics[width=0.82\columnwidth]{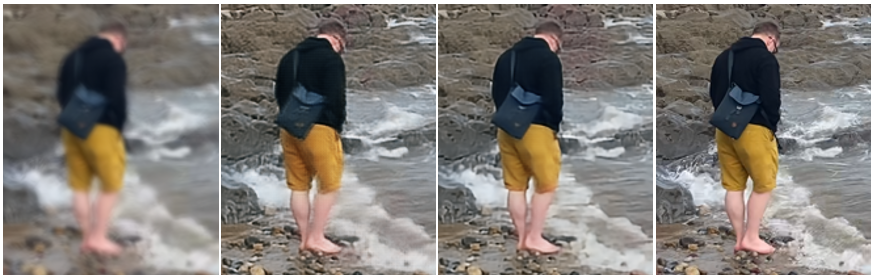}
    \vspace{-2mm}
    \begin{tabular}{x{2.85cm}x{2.85cm}x{2.85cm}x{2.85cm}}
\footnotesize Distorted & \footnotesize  w/o refine & \footnotesize  w refine & \footnotesize  Ground truth\\ 
\end{tabular} 
    \caption{Result comparison of with and without the refinement module. Top-row: Cropped frame 28 of the synthetic sequence `Shore'. Please zoom in for better visualisation.}
    \label{fig:compareRefine}
\end{figure}

The aim of the refinement module is to remove the remaining distortion after the distortion mitigating module. Testing with the synthetic datasets, the refinement module improves the quality of the restored results: PSNR values by up to 1.5 dB, and SSIM values by up to 0.085. The results in Fig. \ref{fig:compareRefine} demonstrate the case that the distortion mitigating module alone leaves some blur and motion artefacts, whilst the refinement module further removes these distortions, leading to sharper results. The refinement module however adds approximately 35\% more computational time.

\subsubsection{Number of input frames}

\begin{figure} [t!]
    \centering
    \includegraphics[width=0.6\columnwidth]{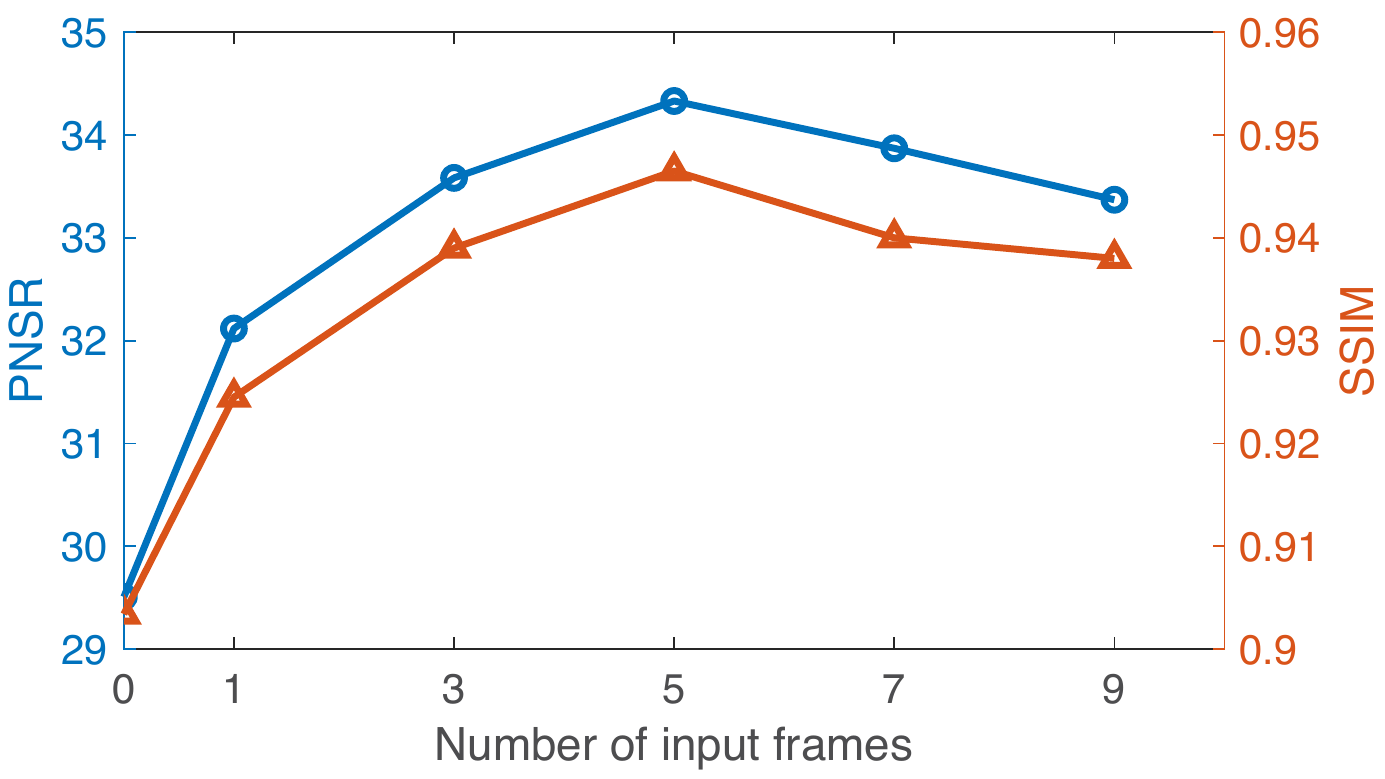}
    \caption{Average quality of restoration using different numbers of input frames. 0 indicates the visual quality of the distorted frames}
    \label{fig:numinputs}
\end{figure}

Several studies have confirmed that removing the atmospheric turbulence effects within a single image is almost impossible, and the higher the number of input frames, the better the restoration \cite{Vint:analysis:2020,Anantrasirichai:Atmospheric:2013}. This is however practical only with static scenes or scenes with static background areas. When moving objects appear in the scene, the imperfect alignment could cause motion artefacts. Our proposed framework employs neither the optical flow estimation nor the wrapping module; therefore, the number of input frames may be limited when fast moving objects are present. Fig. \ref{fig:numinputs} shows the restored qualities of different numbers of input frames ($N_t$, where $N_b$=$N_f$). The average PSNRs and SSIMs indicate that five is the optimal number of frames for our datasets.

\section{Conclusion}
\label{sec:conclusion}

This paper introduces a deep learning-based approach to mitigate atmospheric turbulence distortion in dynamic scenes. Our proposed framework is based on complex-valued convolutions, where amplitudes and phases of local features are fully exploited to remove geometric distortion and to enhance edge and structure information. The framework comprises two modules: distortion mitigating and refinement modules. The models are trained with a combination of synthetic and real datasets. Experiments show that our method performs well even in scenes with strong atmospheric turbulence.

\small
\bibliographystyle{IEEEbib}
\bibliography{ref_heathaze}

\end{document}